\documentclass[letterpaper, 10 pt, journal, twoside]{IEEEtran}
\IEEEoverridecommandlockouts
\usepackage{cite}
\usepackage{amsmath,amssymb,amsfonts}
\usepackage{algorithmic}
\usepackage{graphicx}
\usepackage{textcomp}
\usepackage{xcolor}
\usepackage[caption=false,font=footnotesize]{subfig}
\usepackage{booktabs}
\usepackage{hyperref}
\def\BibTeX{{\rm B\kern-.05em{\sc i\kern-.025em b}\kern-.08em
    T\kern-.1667em\lower.7ex\hbox{E}\kern-.125emX}}
    \hypersetup{
    colorlinks=true,
    linkcolor=blue,
    filecolor=magenta,      
    urlcolor=cyan,
}
\begin{document}

\title{HOPPY: An open-source and low-cost kit for dynamic robotics education}

\author{Joao Ramos$^{1,2}$, 
        Yanran Ding$^{1}$,
        Young-woo Sim$^{1}$,
        Kevin Murphy$^{1}$,
        and Daniel Block$^{2}$
\thanks{Authors are with the $^{1}$Department of Mechanical Science and Engineering and the $^{2}$Department of Electrical \& Computer Engineering at the University of Illinois at Urbana-Champaign, USA. {\tt\footnotesize jlramos@illinois.edu}}}

\maketitle
\pagenumbering{gobble}

\begin{abstract}
This letter introduces HOPPY, an open-source, low-cost, robust, and modular kit for robotics education. The robot dynamically hops around a rotating gantry with a fixed base. The kit lowers the entry barrier for studying dynamic robots and legged locomotion in real systems. The kit bridges the theoretical content of fundamental robotic courses and real dynamic robots by facilitating and guiding the software and hardware integration. This letter describes the topics which can be studied using the kit, lists its components, discusses best practices for implementation, presents results from experiments with the simulator and the real system, and suggests further improvements. A simple controller is described to achieve velocities up to $2m/s$, navigate small objects, and mitigate external disturbances (kicks). HOPPY was utilized as the topic of a semester-long project for the Robot Dynamics and Control course at the University of Illinois at Urbana-Champaign. Students provided an overwhelmingly positive feedback from the hands-on activities during the course and the instructors will continue to improve the kit for upcoming semesters.
\end{abstract}
\begin{IEEEkeywords}
Legged Robots; Education Robotics; Engineering for Robotic Systems
\end{IEEEkeywords}

\section{Introduction}

The imminent robotics revolution will employ robots as ubiquitous tools in our lives. Many machines are already being widely used in factories, assembly lines, and, more recently, in automated warehouses \cite{IEEE_Kiva}. However, most of the tasks performed by these robots are quasi-static, which means that the robot can stop mid-motion without destabilizing (falling down). In contrast, humans can perform dynamic motions in order to complete tasks, like running or weightlifting, more efficiently. In this scenario, the task cannot be interrupted mid-motion. For instance, think how a runner cannot instantaneously freeze motion between steps without falling, or how a Olympic weightlifter cannot statically lift the payload above his/her head. Thus, we must train the next generation of roboticists to create machines that are capable of dynamically performing physical tasks like humans and other animals. Dynamic motions impose unique challenges related to mechanical robustness, actuation saturation, control rates, state estimation quality, and more. However, performing experiments with dynamic robot motions is challenging because capable hardware is expensive and not readily available, and errors can quickly lead to terminal hardware damage \cite{IEEE_RobotsFalling}. To address this issue, researchers created platforms focused on physical robustness, low-cost, and modularity \cite{OpenDynamic,Katz2019}.

Although these dynamic platforms have a significant lower cost than most robots used for research, their use in large-scale education is still impractical. The cost associated with purchasing each unit and the requirement of fabrication tools (CNC machines or high-precision 3D printers), custom electronics, and low-level computing for control and simulation prohibit their use in large cross-disciplinary Engineering courses. In response, this letter introduces HOPPY, a low-cost kit designed for studying dynamic robots. The kit costs under \$500 and is composed of the robot hardware, which can be build exclusively with off-the-shelf components, and a simulator based on the widely adopted software \textit{MATLAB}. The kit covers most of the topics of fundamental robotics courses such as kinematics, dynamics, controls, trajectory generation, simulation, and more. As a tool for nurturing active-learning in robotics, the kit was implemented as a hands-on and semester-long project for the Spring 2020 \textit{Robot Dynamics and Control (ME446)} course at the University of Illinois at Urbana-Champaign (UIUC) (\url{http://youtu.be/6O6czr0YyC8}). This hands-on approach to education is becoming increasingly popular due to its proven effect on enabling enduring, deeper, and more significant learning \cite{EduEng,Hapkit2019}. With this manuscript, we make open-source the robot model, its MATLAB-based simulator, the bill of materials, and assembly instructions. Users can modify and adapt the hardware, software, and simulator as needed for their application and budget. This manuscript and the kit are intended for educators interested in implementing hands-on robotics activities, and for students, researchers, and hobbyists interested in experimentally learning about dynamic robots and legged locomotion. 

\begin{figure}
\centering
    \includegraphics[width=3.4in]{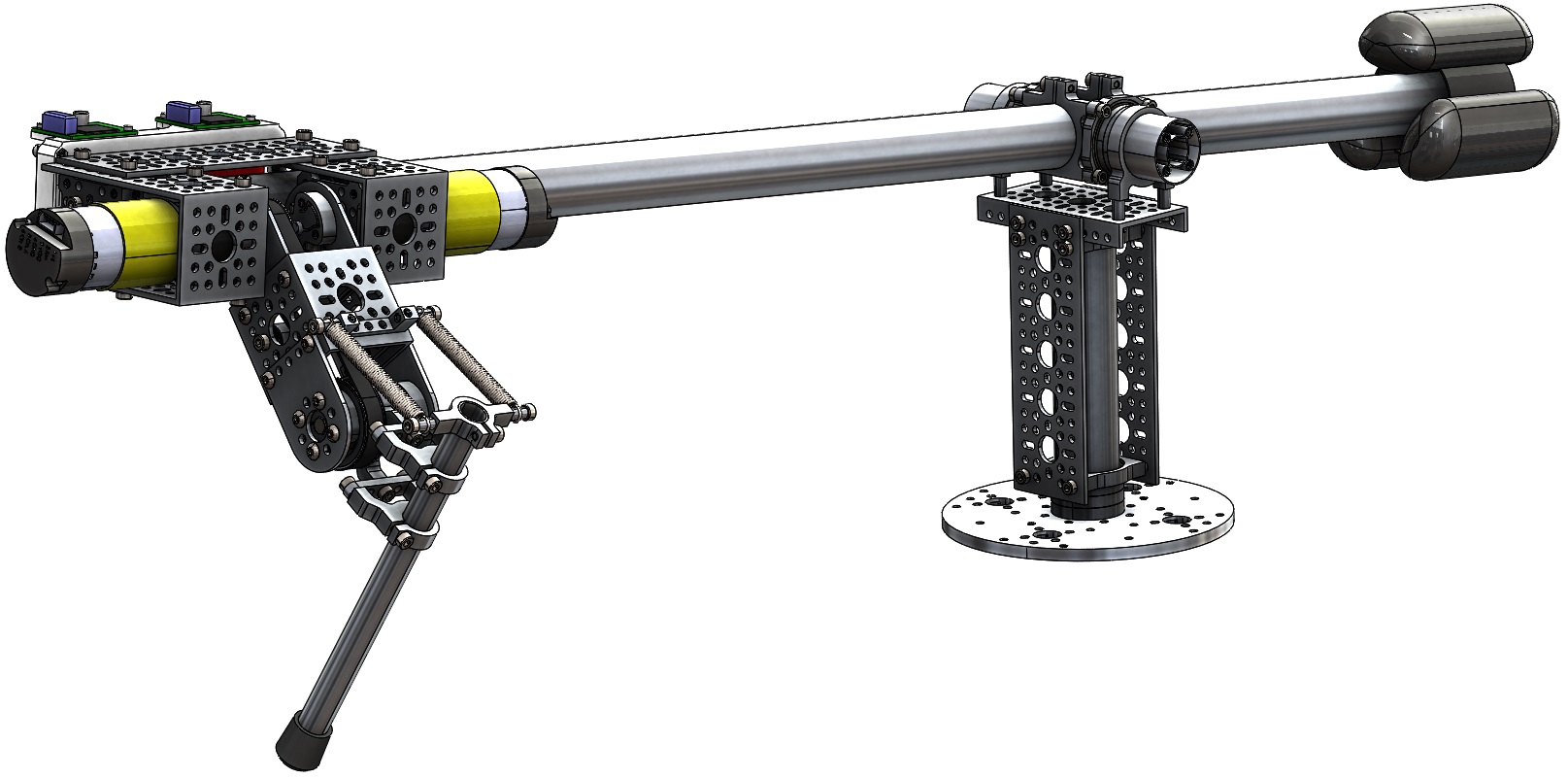}
    \caption{The hopping robot HOPPY for hands-on education in dynamic control and legged locomotion. Available at \url{http:// github.com/RoboDesignLab/HOPPY-Project}.}
    \label{fig_RobotKit}
\end{figure}

This letter is organized as follows. First we describe the topics which are covered by the kit for implementation in fundamental robotics courses. Next, we list the open-source files and instructions made available with this manuscript. We then suggest possible modifications or further improvements to adapt the hardware to the needs of particular educators or researchers. Finally, we discuss our experience implementing the kit in the ME446 course at UIUC and address future work.

\section{Robotics topics covered}
This section describes the (non-exhaustive) list of topics covered in the Robot Dynamics and Control course which are explored using the HOPPY kit. Although here we describe how the kit was modelled and implemented at UIUC, each topic may be explored with more or less detail according to the content of other courses. 

When HOPPY hops around the gantry, it behaves as a \textit{hybrid system} \cite{GrizzleBook}. This means that the dynamics have continuous evolution when the robot is on the ground (stance phase) or airborne (flight phase), but it is interrupted by discrete events when the foot hits the ground (touchdown) or when it takes-off. In this section we describe the equations of motion during stance and flight phases, as well as the transitions between the two. 

\subsection{Kinematics}
\begin{figure}
\centering
    \includegraphics[width=3.4in]{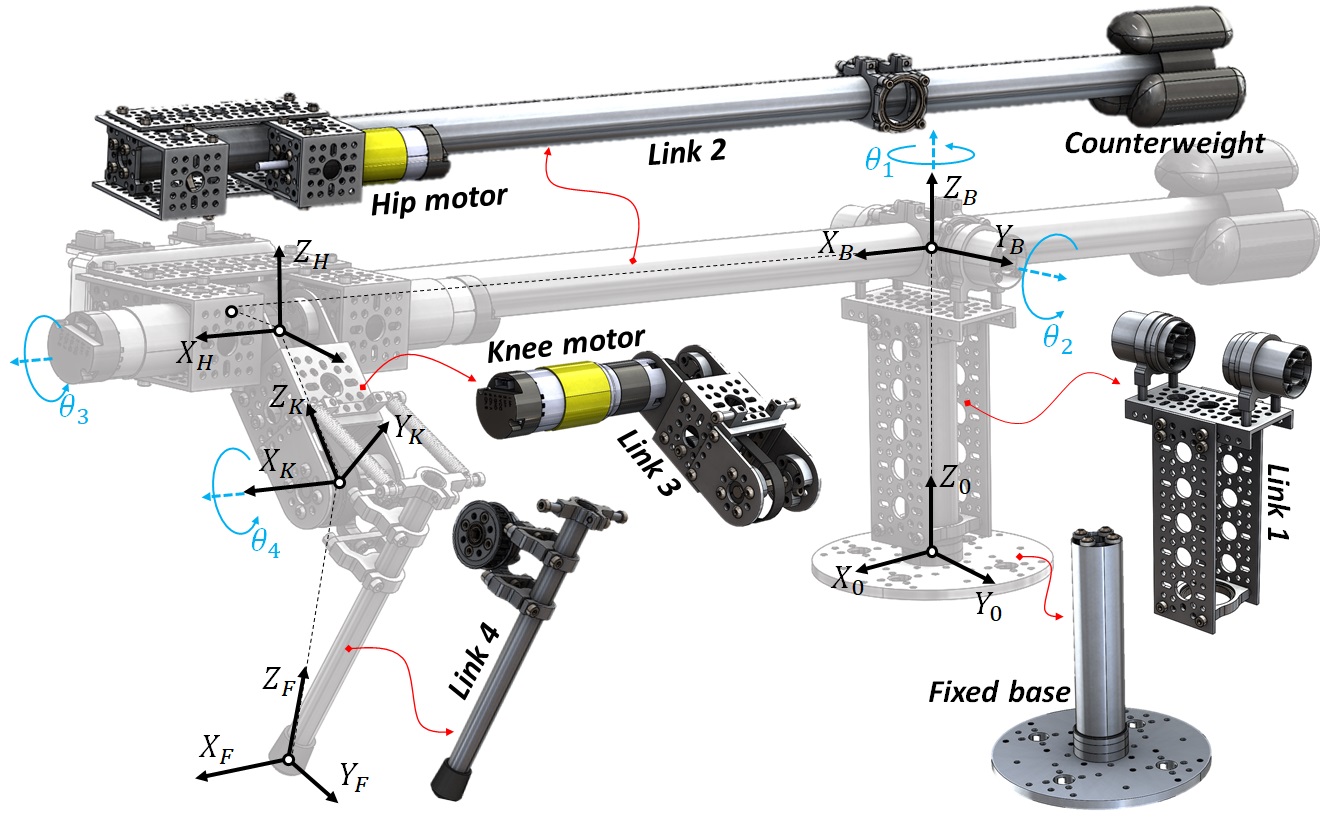}
    \caption{Kinematic transformations and the five rigid-bodies which compose HOPPY. The first two joints ($\theta_1$ and $\theta_2$) are passive while joints three and four are driven by electric motors.}
    \label{fig_Kin}
\end{figure}

The kinematic transformations of HOPPY are shown in fig. \ref{fig_Kin}. The robot consists of a two degrees-of-freedom (DoF) active leg attached to a supporting gantry with two additional DoF's. The origin coordinate frame $X_0Y_0Z_0$ at the base of the rotating gantry is fixed to the ground. Joints $\theta_{1,2}$ are passive (no actuation) while joints $\theta_{3,4}$ are driven by electric motors described in section \ref{sec_Motor}. When hopping, the robot can translate horizontally around $\theta_1$ and vertically around $\theta_2$. If the gantry length $L_B$ is sufficiently long, the robot moves approximately in a 2D (Sagittal) plane \cite{GrizzleBook}.  
The transformation between coordinate frames is described by the Homogeneous Transformation Matrices (HTM) given by $T^k_{k+1}$, which converts a vector $p^{k+1}$ expressed in coordinate frame ${k+1}$ into a vector $p^k$ expressed in coordinate frame $k$  \cite{Spong}:
\begin{equation}
    p^k = T_{k+1}^k p^{k+1} =
    \begin{bmatrix}
         R_{k+1}^k & d^{k}  \\
         0_{1\times3} & 1
    \end{bmatrix}p^{k+1}.
\end{equation}
Where the term $0_{1\times3}$ represents a $1$ by $3$ matrix of zeros. The HTM is composed of a rotation matrix $R_{k+1}^k\in SO(3)$ and a Cartesian translation $d^k\in \mathbb{R}^3$ represented in frame $k$. For instance, the HTM between frames $X_0Y_0Z_0$ and $X_BY_BZ_B$ is given by
\begin{equation}
    T^0_B =
    \begin{bmatrix}
         cos(\theta_1) & -sin(\theta_1) & 0 & 0 \\
         sin(\theta_1) & cos(\theta_1) & 0 & 0 \\
         0 & 0 & 1 & H_B \\
         0 & 0 & 0 & 1 
    \end{bmatrix}.
\end{equation}
Where $H_B$ is the height of the gantry. The position $p_{foot}^0$ of the foot in respect to the base frame expressed in world-fixed coordinates is
\begin{equation}
    \begin{bmatrix}
         p_{foot}^0 \\
         1 
    \end{bmatrix} = T^0_F\begin{bmatrix}
         p_{foot}^F \\
         1 
    \end{bmatrix} = T^0_B T^B_H T^H_K T^K_F\begin{bmatrix}
         0_{3\times1} \\
         1 
    \end{bmatrix}.
\end{equation}
The complete kinematic transformations for HOPPY are described in the provided MATLAB-based simulator.

\subsection{Rigid-body dynamics during flight phase}
The robot state vector is given by the collection of joint angles $q=[\theta_1 \quad \theta_2 \quad \theta_3 \quad \theta_4]^T$ and their time rate of change $\dot{q}$. When the foot is not contacting the ground and the robot is airborne, the equation of motion (EoM) of the robot is given by the well-known \textit{manipulator equation}
\begin{equation}
    M(q)\ddot{q} + C(q,\dot{q})\dot{q} + G(q) = \tau,
    \label{eq_EOM}
\end{equation}
which is derived from energy principles using the system's Lagragian as described in \cite{Spong}. Where $M(q)$ represents the symmetric and positive definite inertia matrix as a function of the configuration, $C(q,\dot{q})$ is the Coriollis and centrifugal forces matrix, $G(q)$ is the vector of torques due to gravity, and $\tau$ is the vector of input torques. The system is underactuated because the first two joints are passive ($\tau_1=\tau_2=0$), and thus, there are four overall DoF but only two inputs \cite{Underactuated}. The input vector is given by
\begin{equation}
    \tau = B_e u + \begin{bmatrix}
         0 \\ 0 \\ 0 \\\tau_s(\theta_4)
    \end{bmatrix} = \begin{bmatrix}
         0 & 0 \\
         0 & 0 \\
         1 & 0 \\
         0 & 1
    \end{bmatrix}\begin{bmatrix}
         \tau_H \\ \tau_K
    \end{bmatrix} + \begin{bmatrix}
         0 \\ 0 \\ 0 \\\tau_s(\theta_4)
    \end{bmatrix},
    \label{eq_tau}
\end{equation}
where $B_e$ is a selection matrix which maps the input torques $u$ to the full system dynamics and $\tau_s(\theta_4)$ represents the torque exerted by the knee spring as a function of the knee angle $\theta_4$. The robot controller can only select torques for the hip $\tau_H$ and knee $\tau_K$ in order to hop around the gantry. The spring attached in parallel with the knee alleviates the joint's torque requirements for hopping. The expression for $\tau_s(\theta_4)$ and the rigid-body parameters for each link mass, center of mass (CoM) position, and inertia tensor are provided in the instructions. 

\subsection{Actuator dynamics model}
\label{sec_Motor}
Equation \eqref{eq_tau} assumes that the actuators are perfect torque sources which can generate arbitrary torque profiles. This assumption may be reasonable for simulation, but in reality the actuator dynamics play a major role in the overall system behavior and stability \cite{ModernRobotics}. Sophisticated drivers can rapidly regulate the current $i$ in the motor coils, which is approximately linearly proportional to the output torque $\tau = k_Ti$ if the steel saturation is neglected ($k_T$ is the motor constant \cite{Katz2019}). However, such drivers are expensive (over \$100) and here we employ the \$25 \textit{Pololu VNH5019} in our kit, which can only control the voltage $V$ across the motor terminals. In this situation, the back-electromotive force (back-EMF) significantly effects the actuator behavior. On the other hand, this limitation provides a valuable opportunity for students to explore how the actuator performance affects the overall dynamics. The model for a brushed and brushless electric motor assuming negligible coil inductance is given by \cite{ModernRobotics}
\begin{align}
    V = R_wi + k_vN\dot{\theta}, \label{eq_Motor_ele}\\
    \ddot{\theta}N^2I_{r} = k_Ti.\label{eq_Motor_mec}
\end{align}
Where $R_w$ is the coil resistance, $k_v$ is the speed constant in $\frac{Vs}{rad}$, $\dot{\theta}$ is the joint velocity (after the gear box), $I_r$ is the motor rotor inertia, and $N$ is the gearbox speed reduction ratio. The hip and knee actuators in HOPPY utlize identical electric motors, but with different gear ratios ($N_H=26.9$ and $N_K=28.8$). Equation \eqref{eq_EOM} is augmented to include the actuator dynamics using
\begin{equation}
    (M(q) + M_{r})\ddot{q} + (C(q,\dot{q}) + B_{EMF})\dot{q} + G(q) = \tau.
    \label{eq_EOM2}
\end{equation}
Where $M_{r}$ is the inertial effect due to rotor inertia and $B_{EMF}$ is the (damping) term due to the back-EMF:
\begin{align}
    M_{r} = I_r\begin{bmatrix}
        0_{2\times2} & 0_{2\times2} \\ 0_{2\times2} & \begin{bmatrix}
        N_H^2 & 0 \\ 0 & N_K^2
    \end{bmatrix}
    \end{bmatrix}, \\
    B_{EMF} = \frac{k_vk_T}{R_w}\begin{bmatrix}
        0_{2\times2} & 0_{2\times2} \\ 0_{2\times2} & \begin{bmatrix}
        N_H^2 & 0 \\ 0 & N_K^2
    \end{bmatrix}
    \end{bmatrix},
\end{align}
and the updated selection matrix and control inputs are given by
\begin{align}
    B_e = \frac{k_T}{R_w}\begin{bmatrix}
        0_{2\times2} \\ \begin{bmatrix}
        N_H & 0 \\ 0 & N_K
    \end{bmatrix}
    \end{bmatrix}, \\
    u = \begin{bmatrix}
        V_H & V_K
    \end{bmatrix}^T,
\end{align}
which illustrates that the input to the system is the voltage applied to the motors, not the torque they produce. Actuators with large gearing ratios have high output inertia (due to the $N^2$ term in \eqref{eq_Motor_mec}) and are usually non-backdrivable because of friction amplification. Thus, we select minimal reduction ratio in HOPPY to mitigate the effect of shock loads in the system during touchdown \cite{Kenneally2016,WensingProprio}. Joint friction can have major (nonlinear) effects on the robot dynamics and models for stiction, dry coulomb friction, viscous friction, and more are suggested in \cite{ModernRobotics} and can be included in the model.  

\subsection{Rigid-body dynamics during stance phase} 
\begin{figure}
\centering
    \includegraphics[width=2in]{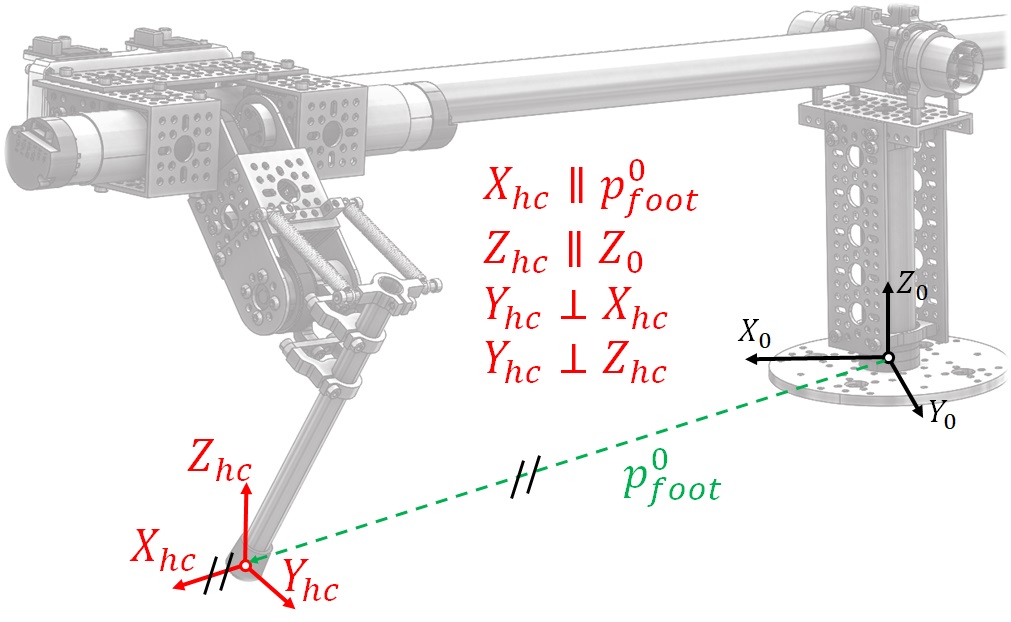}
    \caption{The frame used for the holonomic constraint for non-slip condition. Frame $X_{hc}Y_{hc}Z_{hc}$ changes every time the robot establishes a new foothold and does not necessarily coincides with foot-fixed frame $X_{F}Y_{F}Z_{F}$.}
    \label{fig_Holonomic}
\end{figure}

Equations \eqref{eq_EOM} and \eqref{eq_EOM2} are valid when the robot foot is not touching the ground. When contact occurs, the EoM must describe how the ground reaction forces (GRF) affect the robot motion. Contact mechanics are challenging to simulate in complex systems and commercial physics simulation engines, such as \textit{MuJoCo} \cite{MujocoDyn}, employ different numerical techniques to approximate them. Two common approximations are know as \textit{soft} and \textit{hard} contact models \cite{GrizzleBook}. The soft contact model approximates the contact mechanics by modelling the ground as a spring and damper. Thus, the foot \textit{must} penetrate the ground to generate contact forces. Because the ground is assumed to have high stiffness values, small deformations render large contact forces, and the solver may encounter numerical instability \cite{FeatherstoneBook}. In contrast, the hard contact model, which is used in our simulator, does not allow the foot the to penetrate the ground. It includes a \textit{holonomic} constraint which assumes that the foot does not slip in the $Y_{hc}$ and $Z_{hc}$ directions, shown in fig. \ref{fig_Holonomic}, when in contact with the ground. However, the foot is allowed to slip in the $X_{hc}$ direction, otherwise the system would be overconstrained. The $Z_{hc}$ axis is always vertical, axis $X_{hc}$ is colinear with the line that connects the gantry base and the foot, and $Y_{hc}$ is perpendicular to these. Frame $X_{hc}Y_{hc}Z_{hc}$ changes every time the robot establishes a new foothold and does not necessarily coincides with foot-fixed frame $X_{F}Y_{F}Z_{F}$. 
We utilize the Jacobian $J_{hc}\in \mathbb{R}^{2\times4}$ to write a holonomic constraint for the non-slip condition
\begin{equation}
    J_{hc}\dot{q} = \begin{bmatrix}
    v_{Yhc} \\ v_{Zhc}
    \end{bmatrix} = \begin{bmatrix}
    0 \\ 0
    \end{bmatrix}.
    \label{eq_Holonomic}
\end{equation}
Where $v_{Yhc}$ and $v_{Zhc}$ are the foot linear velocity in the $Y_{hc}$ and $Z_{hc}$ directions. Equation \eqref{eq_Holonomic} is satisfied if \cite{GrizzleBook}
\begin{equation}
    J_{hc}\ddot{q} + \dot{J}_{hc}\dot{q} = 0_{2\times1} 
    \label{eq_dHolonomic}
\end{equation}
is also satisfied. Moreover, when the robot pushes against the ground, it generates reaction forces $F_{GRF} = [F_{Yhc} \quad F_{Zhc}]^T$ which alter the EoM by $J_{hc}^TF_{GRF}$. Finally, the complete stance dynamics is represented by the matrix equation which includes the manipulator equation plus the holonomic constraint:
\begin{equation}
    \begin{bmatrix}
    M(q) & -J_{hc}^T \\
    J_{hc} & 0_{2\times2}
    \end{bmatrix} \begin{bmatrix}
    \ddot{q} \\ F_{GRF}
    \end{bmatrix} = \begin{bmatrix}
    \tau - C(q,\dot{q})\dot{q}-G(q) \\ 
    -\dot{J}_{hc}\dot{q}
    \end{bmatrix}.
    \label{eq_EOM_Holonomic}
\end{equation}
Hence, when solving the forward dynamics, the numeric integrator (MATLAB's \textit{ode45()}) takes the input vector $u$ from the control law and calculates the joints acceleration $\ddot{q}$ and the resultant contact forces $F_{GRF}$ at every iteration.

\subsection{Foot impact model at touchdown}
The impact dynamics are challenging to model realistically and they may cause numerical issues for the solver. Our simulator assumes a completely inelastic collision, which means that the foot comes to a complete stop after hitting the ground. And thus, the robot looses energy every time the foot impacts the ground \cite{WensingProprio}. This assumption implies that the impulsive impact forces remove energy from the system, and the velocity instantaneously changes from $\dot{q}^-$ at time $t^-$ before the impact to $\dot{q}^+$ immediately after the impact at time $t^+$. However, because this discontinuous changes occurs instantaneously, the joint angles remain unchanged ($q^-=q^+$). To compute the robot velocity after the impact we integrate both sides of equation \eqref{eq_EOM_Holonomic} between $t^-$ and $t^+$. Because the joint angles and control inputs do not change during impact we obtain \cite{GrizzleBook,Yanran2017}
\begin{equation}
    M(\dot{q}^+-\dot{q}^-) = J_{hc}^TF_{imp},
\end{equation}
in addition to the constraint due to the inelastic collision $J_{hc}\dot{q}^+ = 0$.
We utilize these equations to write the \textit{impact map} which defines the transition between the aerial phase described by the dynamic EOM \eqref{eq_EOM} to the stance phase described by equation \eqref{eq_EOM_Holonomic} \cite{GrizzleBook}:
\begin{equation}
    \begin{bmatrix}
    \dot{q}^+ \\ F_{imp}
    \end{bmatrix} = \begin{bmatrix}
    M(q) & -J_{hc}^T \\
    J_{hc} & 0_{2\times2}
    \end{bmatrix}^{-1} \begin{bmatrix}
     M(q)\dot{q}^- \\ 0_{2\times1}
    \end{bmatrix}.
    \label{eq_ImpactMap}
\end{equation}
Notice that the impact force $F_{imp}$ is affected by the inverse of the inertia matrix, which is amplified by the reflected inertia of the rotors. Hence, actuators with high gear ratios will create large impact forces, and thus, HOPPY employs minimal gearing \cite{WensingProprio}.

\subsection{Numerical simulation}
\begin{figure}
\centering
    \includegraphics[width=3.4in]{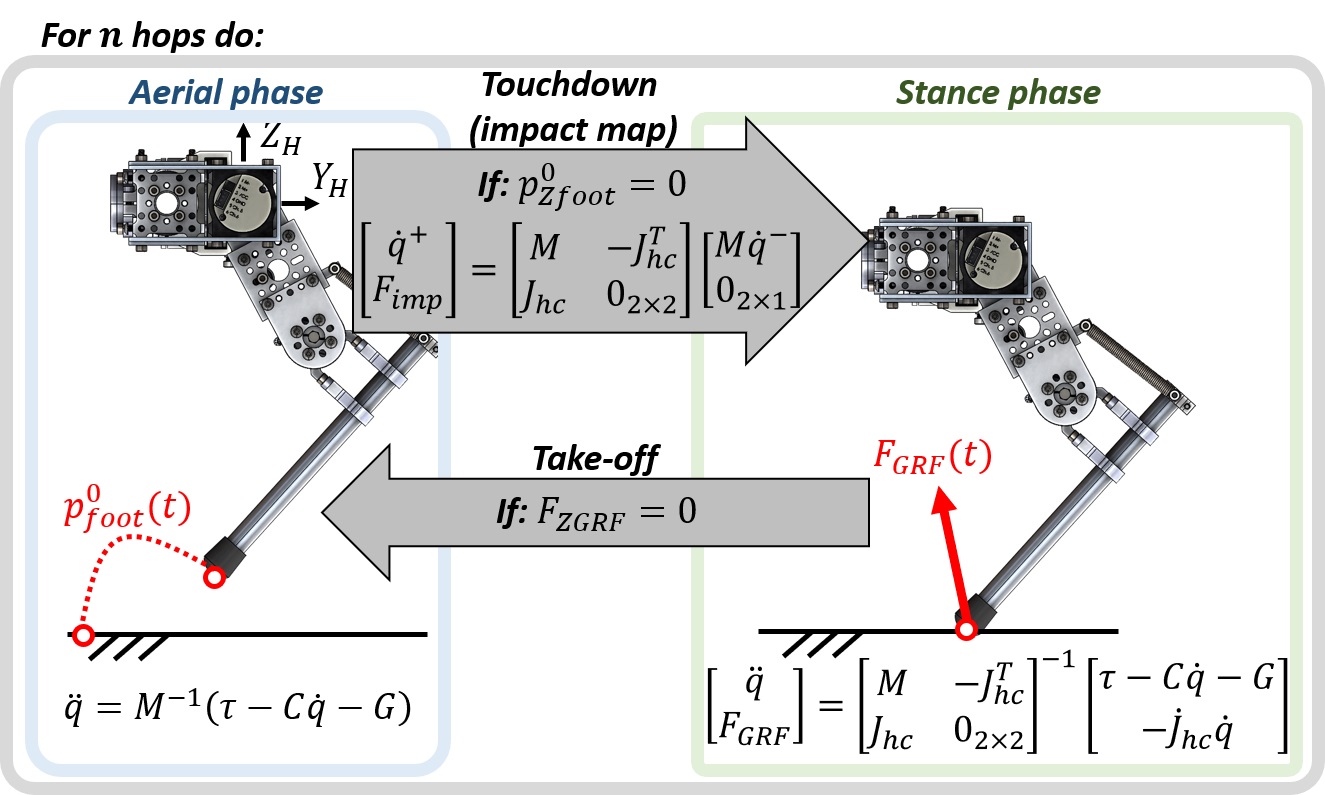}
    \caption{Simulation loop of the hybrid dynamics of HOPPY for $n$ hops. Continuous dynamics in flight (equation \eqref{eq_EOM}) and stance (equation \eqref{eq_EOM_Holonomic}) are interrupted by discrete events: touchdown and lift-off.}
    \label{fig_SimScheme}
\end{figure}

A \textit{hybrid system} is a dynamical system that exhibits both continuous and discrete behavior. HOPPY displays continuous dynamics during flight phase or stance phase, but experiences discontinuous transitions between the two. To simulate hybrid systems, solvers like MATLAB's \textit{ode45()} utilize guard functions to flag the transitions between continuum phases. For instance, the numerical simulation starts with the aerial phase dynamics and is interrupted when the \textit{touchdown} guard function (foot height) cross a certain threshold (ground height), or: $p^0_{Zfoot}(t) = 0$. Similarly, the robot takes-off the ground when the vertical component of the ground reaction force reaches a value equal to zero: $F_{ZGRF}(t)=0$. This sequence is shown in fig. \ref{fig_SimScheme}. The simulator starts in the aerial phase and runs the integration for $n$ hops, which is defined by the number of the transitions between the stance to aerial phase. The simulator is described in detail in the provided instructions. 

\subsection{Hopping control framework}
\label{controls}
A challenging aspect of controlling hopping is that the foot experiences little resistance during aerial phase, but \textit{infinite} resistance when touching the ground during stance phase. Hence, the controller must switch between different policies for the two scenarios. During aerial phase, the foot may follow a predefined trajectory to reach a desired stepping location at impact, as shown in fig. \ref{fig_SimScheme} left. However, during stance the foot pushes against the ground to apply a desired impulse to the CoM. Thus it is a common practice to perform position control of the leg during swing and force control of the ground contact force during stance. We employ a simple strategy in which, during flight, the leg holds a desired configuration and waits for touchdown, while, during stance, the foot applies a predefined force profile against the ground. An insightful and accessible discussion on hopping control is provided in \cite{Raibert_RunningFast, RaibertBook}. 

During flight phase, HOPPY simply holds a predefined leg configuration and waits for touchdown. To achieve this, we employ a simple task-space Proportional plus Derivative (PD) control using:
\begin{equation}
    \tau_{aerial} = J^T_c\left[K_{p}(p_{ref}^H-p_{foot}^H) - K_{d}\dot{p}_{foot}^H\right],
    \label{eq_aerial}
\end{equation}
where  $J_c(q) \in \mathbb{R}^{2\times2}$ is the foot contact point Jacobian, which maps the joint velocities to the linear velocity of the foot in respect to the hip frame $[\dot{y}_{foot}^H \quad \dot{z}_{foot}^H]^T = J_c [\dot{\theta}_H \quad \dot{\theta}_K]^T$. In addition, $K_{p,d}$ are diagonal and positive semi-definite proportional and derivative gain matrices, $p_{ref}^H$ is the desired position of the foot in respect to the hip at touchdown, and $p_{foot}^H$ is its current value. The desired torques $\tau_{aerial}$ are mapped using the motor dynamics to the required input voltages $u_{aerial}$:
\begin{equation}
    u_{aerial} = \begin{bmatrix}
    \frac{R_w}{k_TN_H} & 0 \\
    0 & \frac{R_w}{k_TN_K}
    \end{bmatrix}\tau_{aerial} + k_v\begin{bmatrix}
    N_H\widehat{\dot{\theta}}_H \\ N_K\widehat{\dot{\theta}}_K 
    \end{bmatrix}.
    \label{eq_T2V}
\end{equation}
However, notice the positive feedback sign before $k_v$ and that $\widehat{\dot{\theta}}_{H,K}$ are the \textit{estimated} joint velocities from an observer or filtered derivative. Thus, overestimation of the speed constant $k_v$ or delay in the estimation of the joint velocities may cause oscillations or even instability. For the presented experiments we use a conservative approximation for the speed constant assuming $k_v\approx0$.

\begin{figure}
\centering
    \includegraphics[width=1.8in]{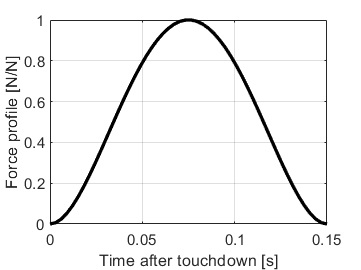}
    \caption{Prescribed force profile for the horizontal ($Y_H$) and vertical ($Z_H$) components of $F_{GRF}$ during a projected stance of $0.15s$. The peak horizontal and vertical forces are modulated to control locomotion speed and hopping height, respectively.}
    \label{fig_Bezier}
\end{figure}

During stance phase, the robot's foot applies a predefined force profile against the ground, which results in an desired net impulse on the robot. The force profile can be generated using simple functions, such as polynomials, in order to be simple to compute online. Here we utilize Bézier polynomials to create a force profile for a projected $0.15s$ stance duration as shown in fig. \ref{fig_Bezier} \cite{Yanran2017}. The force peak is modulated to control jumping height and forward velocity.
To apply the desired force against the ground, we map $F_{dGRF}$ to the desired joint torques using the contact Jacobian similarly to equation \eqref{eq_aerial}:
\begin{equation}
    \tau_{stance} = J_c^TF_{dGRF}.
    \label{eq_TorqueMap}
\end{equation}
Equation \eqref{eq_TorqueMap} assumes that the motion of the lightweight leg does not significantly affect the body dynamics \cite{RaibertBook}. Otherwise, the control law would need to include the torque required to accelerate the leg mechanism using, for instance, Computed Torque Control \cite{Spong}. Torque is mapped to input voltage using equation \eqref{eq_T2V}.

Finally, rapidly switching between the controllers for the stance and aerial phases can cause contact instability issues on the real robot. To smoothly transition between the two we command the input
\begin{equation}
    u = \alpha u_{stance} + (1-\alpha)u_{aerial},
\end{equation}
in which $\alpha$ smoothly changes from $0$ to $1$ after touchdown within a predefined time, typically in the order of $10ms$. 

This proposed control strategy is not affected by the leg singular configuration (straight knee) because it employs the transpose of the contact Jacobian, not its inverse. In addition, the simple strategy also does not require solving the robot inverse kinematics. 

\subsection{Mechatronics}
\begin{figure}
\centering
    \includegraphics[width=3.4in]{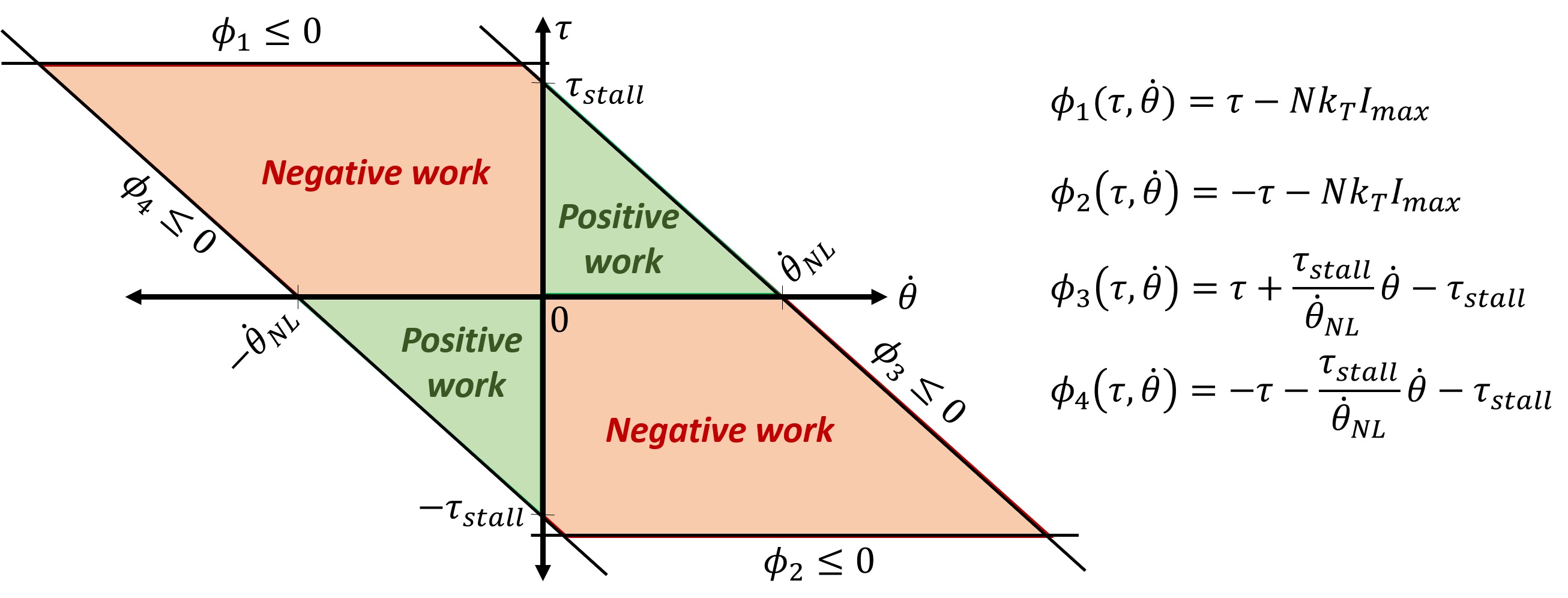}
    \caption{Achievable torque ($\tau$) and speed ($\dot{\theta}$) of the hip and knee joints. Functions $\phi_{1,2}(\tau,\dot{\theta})$ are defined by the gearbox ratio $N$, the torque constant $k_T$, and the maximum current the driver can supply $I_{max}$. Functions $\phi_{3,4}(\tau,\dot{\theta})$ are defined by the voltage supplied $V_{max}$, which limits the stall torque $\tau_{stall} = \frac{V_{max}N}{R_w}k_T$ and no-load speed $\dot{\theta}_{NL}\approx\frac{V_{max}}{N} k_v$ achievable by the joint. The operating region is larger for negative work ($\tau$ and $\dot{\theta}$ with opposite signals) due to the effect of the back-EMF.}
    \label{fig_MotorLim}
\end{figure}

The control of HOPPY involves fundamental concepts related to Mechatronics education. Including:
\begin{itemize}
    \item \textbf{Embedded systems:} The robot controller runs in an onboard microcontroller ($\mu$C) which interfaces with the actuators, sensors, and the host computer. 
    \item \textbf{Discrete control:} The $\mu$C performs computations and interfaces with peripherals at discrete time iterations. And the deterministic execution of these events is fundamental for the implementation of discrete control policies. To effectively control dynamic motions, we target a control rate in the order of $1kHz$ (control loop of $1ms$). 
    \item \textbf{Communication protocols:} The $\mu$C regulates the motor voltage via Pulse Width Modulation (PWM), receives an analog signal which is proportional to the motor current, a binary signal for the foot contact switch sensor, and employs dedicated counters for the incremental encoders.
    \item \textbf{Signal processing:} The encoder has a coarse resolution of 28 counts per revolution (CPR) and the gearbox has noticeable backlash. To avoid noise amplification we estimate the motor velocity using a filtered derivative $\frac{\dot{\theta}(s)}{\theta(s)} = \frac{\lambda s}{\lambda + s}$ converted to discrete time with a period of $T=1ms$, where $s$ is the Laplace variable for frequency, and $\lambda$ is a tunable constant (usually $\lambda\approx10$). Fast sampling rates are essential to avoid delayed velocity estimation. The analog signal from the motor current estimation is also noisy and requires filtering if used for controls.
    \item \textbf{Saturation:} The achievable joint torques and speeds are limited by the available voltage supply ($V_{max}=12V$) and the maximum current that the driver can provide ($I_{max}=30A$) \cite{ModernRobotics}. The operating region of the motors is depicted in fig. \ref{fig_MotorLim}.
\end{itemize}

\section{Kit design and components}
HOPPY is designed exclusively with off-the-shelf components to lower its cost and enable modularity, customization, and facilitate maintenance. In addition, it is lightweight (total weight about $3.8kg$), portable, and mechanically robust. All load-bearing components are made of metal parts from \url{https://gobilda.com/}. We intentionally avoid the use of plastic 3D printed parts due to their short working life under impact loads. Because of the kit's durability, only the initial cost for setting up the course is required and the equipment can be reutilized in future semesters with minimal maintenance costs. Both motors are placed near the hip joint to reduce the moving inertia of the leg and enable fast swing motions. The second actuator is mounted coaxially with the hip joint and drives the knee joint through a timing belt. The motors have minimal gearing ratio in order to be backdrivable. To reduce the torque requirements for hopping, springs are added in parallel with the knee and a counterweight is attached at the opposite end of the gantry as shown in fig. \ref{fig_RobotKit}. We employ cheap $2.3kg$ ($5lb$) weights typically utilized for gymnastics. The addition of an excessively heavy counterweight reduces the achievable frictional forces between the foot and the ground, and the robot is more likely to slip.

\begin{figure}
\centering
    \includegraphics[width=2.8in]{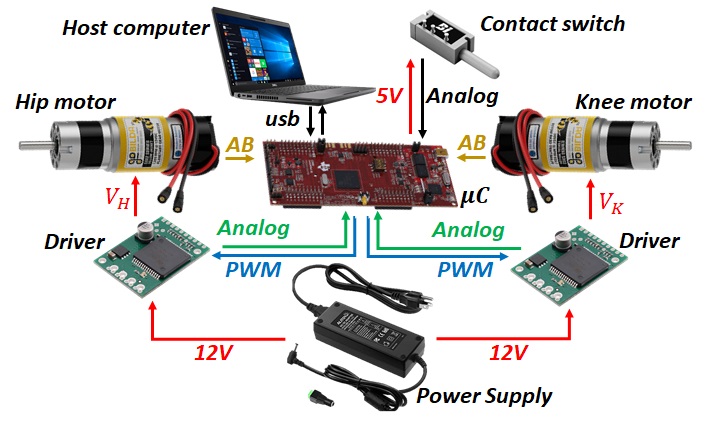}
    \caption{Electrical components of the kit and communication diagram.}
    \label{fig_Electrical}
\end{figure}

The diagram of electrical connections and communications protocols is depicted in fig. \ref{fig_Electrical}. The user computer communicates with the embedded $\mu$C (\textit{Texas Instruments LAUNCHXL-F28379D}) via a usb port. The $\mu$C commands the desired voltage to the motors (\textit{goBilda 5202-0002-0027} for the hip and \textit{0019} for the knee) via PWM to the drivers (\textit{Pololu VNH5019}), and receives analog signals proportional to the motor current and the pulse counts from the incremental encoders. The motor drivers are powered by a 12V power supply (\textit{ALITOVE 12Volt 10 Amp 120W}). A linear potentiometer with spring return (\textit{TT Electronics 404R1KL1.0)} attached in parallel with the foot works as a foot contact switch.   

The kit available at \url{https://github.com/RoboDesignLab/HOPPY-Project} is composed of:
\begin{enumerate}
    \item the solid model in \textit{SolidWorks} and .STEP formats; 
    \item the video instructions for mechanical assembling; 
    \item the electrical wiring diagram
    \item the dynamic simulator code and instructions; 
    \item the complete list of components and quantities; and 
    \item the basic code for the microcontroller.
\end{enumerate}

\section{Modifications, limitations, and safety}
\label{sec_Modifications}
The kit is designed to facilitate the adjustment of physical parameters and replacement of components. For instance, gearboxes with ratios of $3.7:1$ to $188:1$ can be used in either the hip or knee. The length of the lower leg is continually tunable between $60mm$ and $260mm$. Additional springs can be added (or removed) to the knee joint if necessary. The gantry length is continually adjustable between $0.1m$ and $1.1m$ and its height has multiple discrete options between $0.2m$ and $1.1m$. The dynamic effect of the counterweight can be adjusted by shifting its position further away from the gantry joint or adding more mass.

However, the kit also presents performance limitations. For instance, voltage control of the motor is not ideal as it hinders the precise regulation of ground reaction forces at high speed. The coarse resolution of the encoders and the backlash in the gearbox also degrades position tracking performance. The hobby-grade brushed motors have limited torque density (peak torque capability divided by unit mass), and thus the robot \textit{will not} be able to hop without the aid of the springs and the counterweight. In addition, the simulator assumes simple dynamic models for impact and contact, and neglects friction, structural compliance, vibrations, and foot slip. Due to these limitations, it is \textit{unlikely} that the physical robot will behave exactly like the simulator predicts. However, the simulator is a valuable tool to obtain insights about the behavior of the real robot, it can be used for preliminary tuning of controllers before implementation in the hardware, and is a fundamental tool for teaching.

When experimenting with HOPPY (or any dynamic robot) some safety guidelines must be followed. Students should wear safety goggles and clear the robot's path. It is advised to \textit{constantly} check the temperature of the motor's armature. If they are too hot to be touched, the experiments should be interrupted until cool down. Aggressive operations (high torque and speeds) within negative work regimes should be done carefully because electric motors regenerate part of the input energy back into the power supply when backdriven \cite{Seok2013}. And thus, utilizing a battery is more appropriated if considerable negative work is expected. For instance, a conventional $12V$ car battery can be used both as power supply and a mechanical counterweight. Finally, it is advised to often check for loose components due to the constant impacts and vibrations. 

If the budget allows, some components of HOPPY can be improved. For instance, (i) employing encoders with finer resolution around 12bits (4096 counts per revolution) would be ideal. This would improve position control and joint velocity estimation. (ii) Implementation of motor drivers which can perform high-speed current control, such as the \textit{Advanced Motion Control AZBDC12A8}. This feature would enable precise torque control and considerably improve force tracking during stance. An even better solution would be (iii) the utilization of brushless motors similar to those employed in \cite{OpenDynamic,Katz2019}. These have much superior torque density and would enable hopping without the aid of springs or counterweights. (iv) Additional sensors (encoders) could be added to the gantry joints ($\theta_{1,2}$) or an Inertial Measurement Unit (IMU) can be mounted on the robot for state estimation \cite{CassieStateEstimation}. (v) Data logging can be achieved using external loggers such as the \textit{SparkFun OpenLog}. And finally, (vi) to enable unrestricted rotations around the gantry, a battery can be used to power the robot and a USB slip ring can be included for communication between the $\mu$C and the host's computer.  

\section{Results and Discussion}
\begin{figure*}
\centering
    \includegraphics[width=7in]{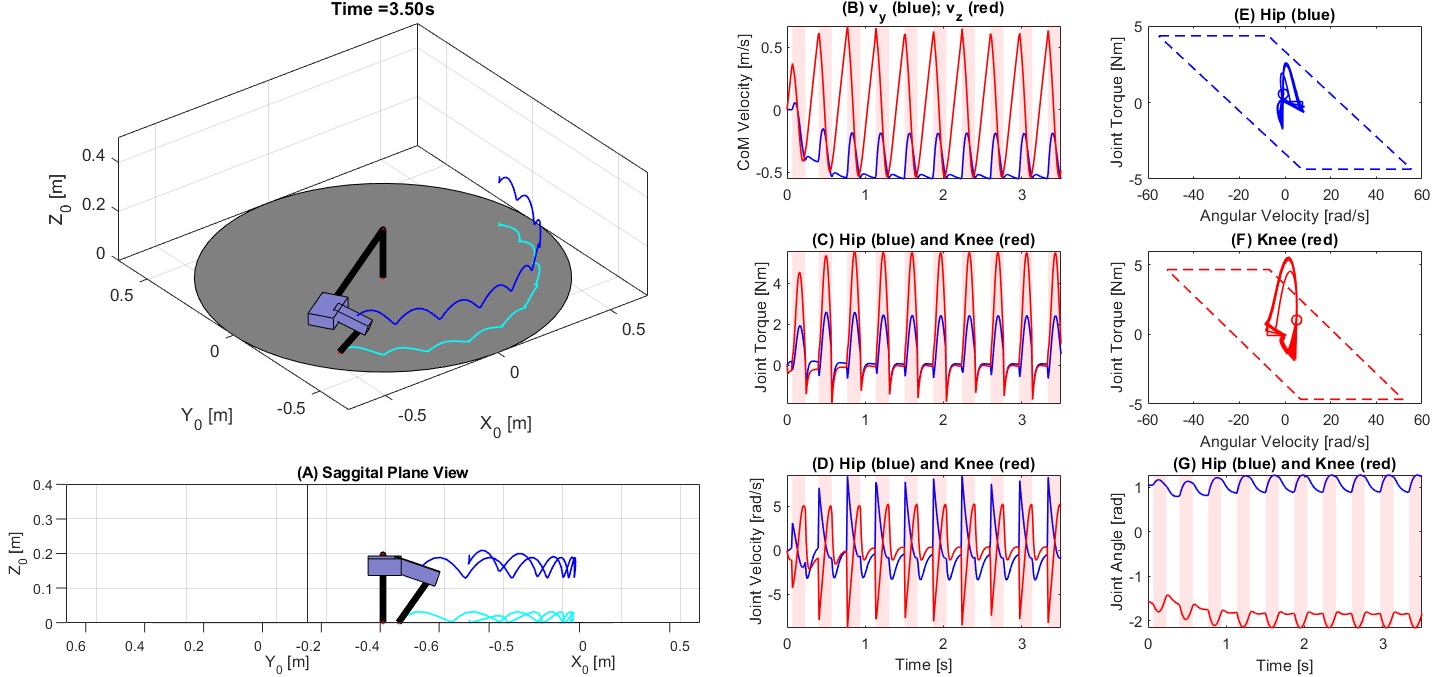}
    \caption{Screenshot of the output screen of the simulator. It shows the robot animation in isometric and Sagittal plane views along with the trajectories of the hip joint in blue and the foot in cyan. Plot (B) shows the horizontal and vertical tangential velocities of the hip base around the gantry. Areas shaded in red represent stance phases. Hip and knee joint torques are shown in (D) and their angular velocities in (D). Plots (E) and (F) display the limiting polygon for the hip and knee actuators along with the torque versus velocity trajectory. Feasible trajectories must be always contained inside the polygon. Finally, plot (G) shows the hip an knee joint angles over time.}
    \label{fig_SimScreenshot}
\end{figure*}
A snapshot of the output of the MATLAB-based simulator is shown in fig. \ref{fig_SimScreenshot}. It displays an animation of the robot during hopping (including a view of the robot's Sagittal plane) with a trace of the hip joint and foot spatial trajectories. The visualizer also shows the hopping velocity, the joint states, required torques, and the working regions of the actuators to check for saturation. Areas shaded in red represent stance phases. More information about the simulator is provided in the accompanying instructions.

\begin{figure*}
\centering
    \includegraphics[width=7in]{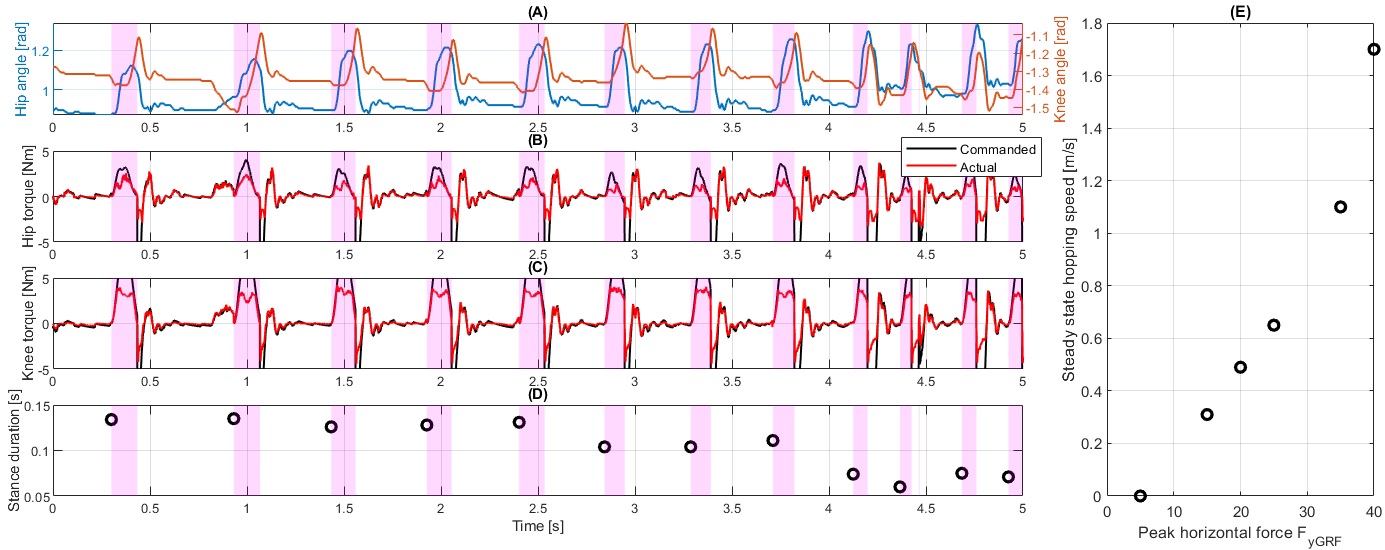}
    \caption{Experimental data with HOPPY speeding up from around $0.65m/s$ to $1.2m/s$. Shaded magenta areas represent stance periods. (A) Hip ($\theta_3$) joint angles in the left vertical axis and knee ($\theta_4$) angle in right axis. (B) Hip joint commanded torque (black) and actual torque (red) estimated from the measured current. (C) Knee joint torque. (D) Stance phase duration (duration of following shaded areas). And (E) steady-state hopping speed as a function of the commanded peak horizontal contact force $F_{yGFP}$.}
    \label{fig_Experiment_Data}
\end{figure*}

For the experiments in companion video (\url{https://youtu.be/_lbKIpiRWKI}), we employed the controller described in section \ref{controls} using aerial phase gains of $K_p = 500[1_{2\times2}]$ and $K_s = 50[1_{2\times2}]$, the stance duration $T_s=0.15s$, and the peak vertical force commanded of $80N$. Where $1_{2\times2}$ is a $2\times2$ identity matrix. The peak horizontal force was modulated to control the locomotion speed. Fig. \ref{fig_Experiment_Data}-(A) to (D) shows the data of robot accelerating from about $0.65m/s$ to around $1.2m/s$ by increasing the horizontal force peak. The joint torques are estimated using the measured motor current ($\tau=ik_TN$). The saturation occurs when the requested input voltage is above the supplied $12V$. Fig. \ref{fig_Experiment_Data}-(D) shows the stance duration, measured from the contact switch, which decreases according to the locomotion speed. Notice that at fast speeds, the robot touches the ground for less that $100ms$. Finally, \ref{fig_Experiment_Data}-(E) shows the achieved steady-state hopping velocities according to the commanded peak horizontal force. We note that the locomotion speed is likely overestimated because it was computed using the relative velocity between the foot and the hip during stance, assuming the the foot does not slip during contact.   

\begin{figure*}
\centering
    \includegraphics[width=7in]{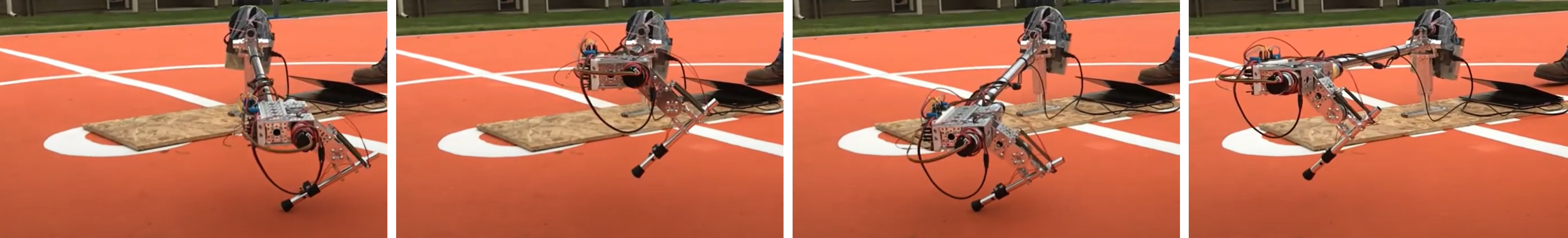}
    \caption{Snapshots of the hopping experiment with HOPPY by one of the student teams (\url{https://youtu.be/6O6czr0YyC8}).}
    \label{fig_Experiment}
\end{figure*}

HOPPY was used in a semester-long project in the \textit{ME446 Robot Dynamics and Control} course at UIUC. The course is attended by senior undergraduate and first-year graduate students from the Departments of Mechanical Science and Engineering, Electrical and Computer Engineering, and Industrial \& Enterprise Systems Engineering. Ten cross-departmental teams of four to five students each received their own leg (hip base of link 2 plus links 3 and 4), and all teams shared six gantries (fixed base plus links 1 and 2). During the first half of the course, students derived the kinematic and dynamic model of the robot, developed different controllers, and created a simple version of the robot simulator in MATLAB. They assembled their kits half-way through the semester and performed experiments with the robot on the second half. After developing their own dynamic simulator, students were provided with our code for comparison. The results from experiments with the real robot from one of the teams are shown in the snapshots of fig. \ref{fig_Experiment}. The team developed a custom foot contact sensor to detect stance phase. The robot was able to regulate hopping speed, overcome small obstacles, and mitigate disturbances from gentle kicks. 


\section{Conclusion and future work}
This letter introduced HOPPY, an open-source, low-cost, robust, and modular kit for robotics education. The kit lowers the barrier for studying dynamic robot behaviors and legged locomotion with real systems. The control of dynamic motions present unique challenges to the robot software and hardware, and these are often overlooked in conventional robotics courses, even those with hands-on activities. Here we describe the topics which can be explored using the kit, list its components, discuss best practices for implementation, and suggest further improvements. HOPPY was utilized as the topic of a semester-long project for an first year graduate-level course at UIUC. Students provided an overwhelmingly positive feedback from the hands-on activities during the course. The instructors will continue to improve the kit and course content for upcoming semesters. To nurture active learning, future activities will include a friendly competition between teams to elect the fastest robot and the most energy-efficient robot. Speed is estimated by timing complete laps around the gantry. Energy efficiency is determined by the Cost of Transport, which is proportional to the ratio between the total electrical power consumed during one hopping cycle and the average translation speed \cite{Seok2013}. Further improvements of the kit will be made open-source.


\bibliographystyle{plain}
\bibliography{References_Ramos}
\end{document}